\documentclass{article}

\usepackage{CJKutf8}
\usepackage{longcat_style}
\usepackage{adjustbox}
\usepackage[utf8]{inputenc} %
\usepackage[T1]{fontenc}    %
\usepackage{newunicodechar}
\usepackage{hyperref}       %
\usepackage{xcolor}
\usepackage[normalem]{ulem} %
\hypersetup{
    colorlinks=true,      %
    linkcolor=blue,      %
    urlcolor=blue,       %
    citecolor=blue,      %
    linkbordercolor=blue, %
    urlbordercolor=blue,
    citebordercolor=blue,
    pdfborderstyle={/S/U/W 1}, %
}
\usepackage{float}
\usepackage{url}            %
\usepackage{booktabs}       %
\usepackage{amsfonts}       %
\usepackage{nicefrac}       %
\usepackage{microtype}      %
\usepackage{lipsum}		%
\usepackage{graphicx}
\usepackage{natbib}
\usepackage{doi}
\usepackage{amsmath}
\usepackage{amssymb} %
\usepackage{xspace}
\usepackage{enumitem}
\usepackage{multirow}
\usepackage{subcaption} 
\usepackage{makecell}
\usepackage{hyperref, cleveref}
\usepackage{pifont}
\usepackage[inkscapelatex=false]{svg}
\usepackage{caption}
\DeclareCaptionLabelSeparator{pipe}{ | }
\usepackage{tcolorbox}
\newtcolorbox{coloredquote}[1][]{
    colback=green!5!white,  
    colframe=green!70!black, 
    boxrule=2pt,
    arc=7pt,
    left=6pt,
    right=6pt,
    top=4pt,
    bottom=4pt,
    title=#1
}

\setlist[itemize]{leftmargin=*}
\setlist[enumerate]{leftmargin=*}
\setlist[description]{leftmargin=*}

\newcommand{\longcat}{LongCat-Flash-Exp\xspace}

\newcommand{\zigzag}{\includegraphics[width=0.027\textwidth]{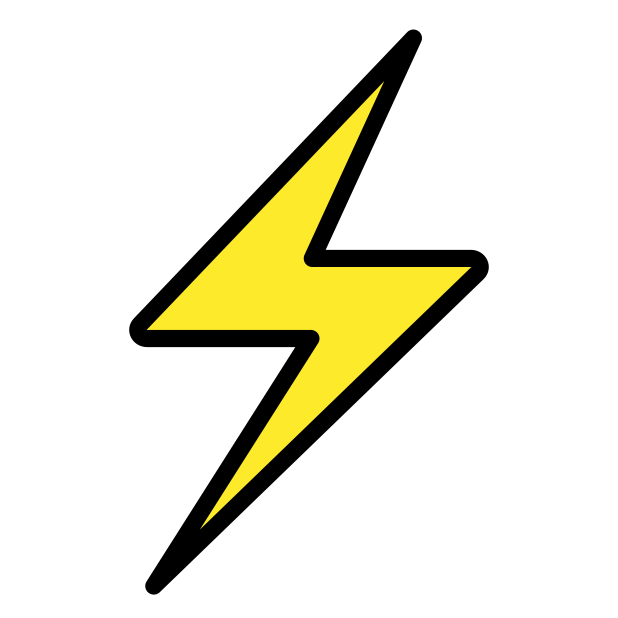}}

\title{Efficient Context Scaling\\ with LongCat ZigZag Attention~\zigzag}

\author{
Chen Zhang, Yang Bai, Jiahuan Li, Anchun Gui, Keheng Wang, Feifan Liu \\
\textbf{Guanyu Wu, Yuwei Jiang, Defei Bu, Li Wei, Haihang Jing, Hongyin Tang, Xin Chen} \\
\textbf{Xiangzhou Huang, Fengcun Li, Rongxiang Weng\footnotemark[1], Yulei Qian, Yifan Lu, Yerui Sun} \\
\textbf{Jingang Wang\thanks{Rongxiang and Jingang are the corresponding authors.}, Yuchen Xie, Xunliang Cai} \\
    Meituan, China \\
	\texttt{zhangchen76@meituan.com} \\
}

\renewcommand{\headeright}{\raisebox{-0.2\height}{\includegraphics[height=1.8em]{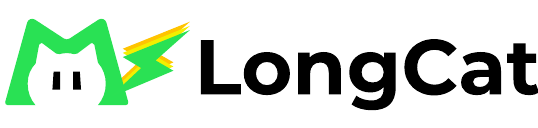}}}
\renewcommand{\shorttitle}{LongCat ZigZag Attention}

\begin{document}
\maketitle
\setcounter{footnote}{0}

\begin{abstract}
We introduce LongCat ZigZag Attention (LoZA), which is a sparse attention scheme designed to transform any existing full-attention models into sparse versions with rather limited compute budget.
In long-context scenarios, LoZA can achieve significant speed-ups both for prefill-intensive (e.g., retrieval-augmented generation) and decode-intensive (e.g.,
tool-integrated reasoning) cases.
Specifically, by applying LoZA to LongCat-Flash during mid-training, we serve LongCat-Flash-Exp as a long-context foundation model that can swiftly process up to 1 million tokens, enabling efficient long-term reasoning and long-horizon agentic capabilities.
\end{abstract}


\begin{figure}[ht]
    \centering
    \includegraphics[width=0.9\textwidth]{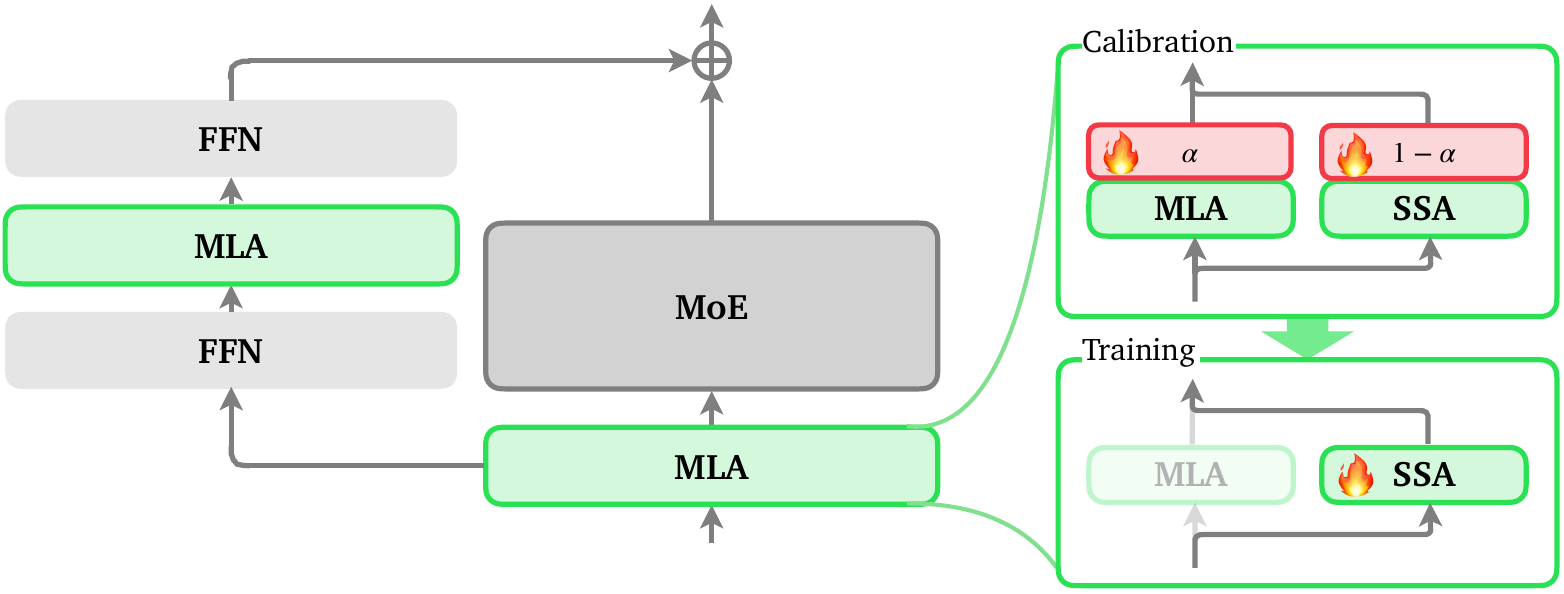}
    \caption{The illustration of LongCat ZigZag Attention (LoZA), which involves first calibration and then training for realizing the sparsity. The illustration is shown with the exemplar shortcuted-MoE~\citep{DBLP:conf/icml/CaiJQC0025} in LongCat-Flash, which encompasses two MLA layers~\citep{DBLP:journals/corr/abs-2405-04434}. SSA: streaming sparse attention~\citep{DBLP:conf/iclr/XiaoTCHL24}.}
    \label{fig:loza}
\end{figure}

\section{Architecture}

Built upon any language models (LMs, e.g., LongCat-Flash~\citep{DBLP:journals/corr/abs-2509-01322}), we can basically alter some of full-attention modules to their sparse-attention alternatives~\citep{DBLP:conf/icml/TangZZXKH24,DBLP:journals/corr/abs-2408-07092,DBLP:conf/icml/ZhangXHWX0C25,DBLP:journals/corr/abs-2506-04108,DBLP:journals/corr/abs-2509-24663,deepseekai2025}. The transition from full attention to sparse attention is usually exerted during mid-training or so~\citep{DBLP:journals/corr/abs-2510-14865,zhang2025interplay,DBLP:journals/corr/abs-2510-23081,DBLP:journals/corr/abs-2508-12407}, and the sparse attention outcome is later used in follow-up procedures.

\paragraph{Full Attention}

Attention~\citep{DBLP:conf/nips/VaswaniSPUJGKP17}, or typically full attention, is a key ingredient in modern transformer architectures and thus large language models (LLMs). Usually, the attention takes the softmax form as below:
\begin{equation}
    \mathbf{O}=\text{softmax}(\mathbf{Q}\mathbf{K})\mathbf{V},
\end{equation}
where other details such like the denominator are omitted. 

As noted in the form, the compute essentially scales quadratically with the increase of context, incurring critical burdens for applications such as retrieval-augmented generation~\citep{DBLP:conf/naacl/ShiMYS0LZY24,DBLP:conf/acl/YenG024} or tool-integrated reasoning~\citep{DBLP:journals/csur/QinHLCDCZZHXHFSWQTZLSXZ25,DBLP:conf/iclr/GouSGSYHDC24} that runs with extensively long input or output.

\paragraph{LongCat ZigZag Attention}

Pioneers have devoted a lot of efforts in exploring how sparse attention could serve as an alternative as below:
\begin{equation}
    \mathbf{O}^{\prime}=\text{softmax}(\mathbf{Q}\mathbf{K}^{\prime})\mathbf{V}^{\prime},
\end{equation}
where notations that are marked with $^{\prime}$ are sparsified ones. For instance, $\mathbf{K}^{\prime}$ would only retain 10\% elements compared to those indicated in $\mathbf{K}$.

In the alternative, the compute minimally remains constant with respect to the context scaling. To echo this, we put forward LongCat ZigZag Attention (short-termed LoZA), as shown in Figure~\ref{fig:loza}. LoZA firstly uncovers the layers that can be sparsified without hurting much performance~\citep{DBLP:conf/iclr/XiaoTZGYTF025}, secondly sparsifies the layers that can be further trained~\citep{deepseekai2025} to close performance gap. The whole process behaves very much like what has been described in \emph{lottery tickets hypothesis}~\citep{DBLP:conf/iclr/FrankleC19}. In theory, a mid-trained LM is sequentially sparsified, rewound, mid-trained to maximally recover the full performance. In other words, the calibration starts at the end of mid-training while the training starts at the beginning of the mid-training. 

\textit{Calibration}. Regarding the MLA~\citep{DBLP:journals/corr/abs-2405-04434} that has been utilized in LMs such as DeepSeek-V3 and LongCat-Flash, LoZA assumes there are totally $n$ MLAs. LoZA initially attaches a unique parameterized factor $a_{i}\in[0, 1]$ per MLA so that the MLA processes as below:
\begin{equation}
    \hat{\mathbf{O}}_{i}=\alpha_{i}\cdot\mathbf{O}_{i}+(1-\alpha_{i})\cdot\mathbf{O}^{\prime}_{i},
\end{equation}
where $\mathbf{O}_{i}$ and $\mathbf{O}^{\prime}_{i}$ denote the full attention output and sparse attention output generated by the $i$-th MLA separately. Here, sparse attention follows the streaming sparse pattern, where one query token only attends to several sink and local blocks~\citep{DBLP:conf/iclr/XiaoTCHL24}.

Then a round of training on the calibration data is carried out by freezing any parameters within the mid-trained LM except all $a_{i}$. After the optimization, $a_{i}$ would differ from each other in magnitude, which is used to signify the importance of the corresponding MLA. Notably, by partly sparsifying MLAs with the lowest $a_{i}$ values, the performance of the LM is largely preserved.

Based on the observation unearthed in calibration, LoZA later turns 50\% MLAs with the lowest $a_{i}$ in the mid-trained LM from full attention to streaming sparse attention (SSA) such that:
\begin{equation}
    \mathbf{O}^{*}=\text{softmax}(\mathbf{Q}\mathbf{K}^{*})\mathbf{V}^{*},
\end{equation}
where $\mathbf{K}^{*}$ and $\mathbf{V}^{*}$ are anchored and blocked keys and values (\#sink blocks $s$, \#local blocks $l$, block size $b$).

\textit{Training}. Preferably, though the sparsified LM maintains competitive performance, training is further required to close the potential performance gap brought by the sparsification, especially in long-context scenarios.

For budget purpose, we decide to locate the training at mid-training. Since mid-training only consumes hundreds of billions of tokens, it is relatively acceptable for limited compute.

\textit{Pilot Studies}. To elaborate how the sparse pattern and training work, we have preliminarily conducted a study on directly applying an interleaved sparse pattern (i.e., sparsifying one out of two adjacent layers), and then tuning the pattern with only a few tokens. 

\begin{table}[ht]
    \centering
    \caption{The pilot studies on calibration and training. The \textit{interleaved sparse pattern} denotes sparsifying one out of two adjacent layers, and the \textit{calibrated sparse pattern} denotes sparsifying the lowest-valued layers during calibration. The \textit{sparse training} means further training after sparsification. LongEval~\citep{longchat2023} is long-context evaluation and others are short-context evaluation.}
    \begin{tabular}{lcccc}
    \toprule
        \textbf{Method} & \textbf{BBH} & \textbf{GSM8K} & \textbf{HumanEval+} & \textbf{LongEval} \\
    \midrule
        LongCat-Flash & 81.0 & 94.2 & 66.6 & 95.7 \\
        \quad w/ interleaved sparse pattern & 81.2 & 94.3 & 65.9 & 54.1 \\
        \qquad w/ sparse training & 80.2 & 93.4 & 70.1 & 67.4 \\
        \quad w/ calibrated sparse pattern & 80.9 & 93.8 & 67.7 & 89.6 \\
    \bottomrule
    \end{tabular}
    \label{tab:pilot}
\end{table}

It shall be observed in Table~\ref{tab:pilot} that 1) hand-crafted sparse pattern yields significant performance drop on long-context data while calibration makes the sparse pattern way better, and 2) training enhances performance on long-context data.

\paragraph{Desiderata}

Our design is to a great extent inspired by Duo-Attention~\citep{DBLP:conf/iclr/XiaoTZGYTF025}. The reason why we instead bet on layer-level rather than head-level streaming sparse attention as in Duo-Attention is that head-level sparsity in our case would easily lead to both compute imbalance across parallel ranks and schedule recompute across attention layers during inference.

\textit{Kernel}. Head-level sparsity complicates kernel control flow, consuming critical efforts in achieving balanced metadata. In contrast, layer-level sparsity allows the kernel to follow a single uniform schedule, minimizing metadata pressure. Furthermore, it is possible that kernels can process multiple KV groups per thread block to maximize occupancy; consequently, distinct head-level sparse patterns could induce warp divergence.

\textit{Engine}. Head-level sparsity can shard heterogeneous workloads to different ranks (e.g., one device processing all full heads while another processes all sparse ones), creating stragglers that bottleneck global synchronization. Differently, layer-level sparsity guarantees uniform compute across all ranks. Besides, layer-level sparse patterns eliminate the runtime overhead of recomputing schedule metadata across layers.

\section{Training}

The training covers mid-training (specifically only long-context extension phases) and follow-up post-training, and finally yields \longcat. The mid-training recipe is roughly the same as those leveraged by LongCat-Flash~\citep{DBLP:journals/corr/abs-2509-01322}. 

Contrarily, the post-training recipe is primarily simplified for fast prototyping. And in fact, we could already achieve expected performance with this simple recipe. We leave more fancy post-training to catch even more fascinating performance as future work. It is noteworthy the post-training recipe is mainly utilized for tuning an instruct (or say chat) model. 

Nonetheless, to unlock the ability of handling a longer context, we equip these recipes with YaRN~\citep{DBLP:conf/iclr/PengQFS24} so that \longcat can extrapolate itself to processing up to 1M tokens. In addition to that, we provide a few crucial parameters involved in LoZA. The block size (i.e., $b$) is 128, the number of sink blocks (i.e., $s$) is 1, and the number of local blocks (i.e., $l$) is 7, summing to 1,024 tokens. 

\paragraph{Mid-training} For mid-training, we take a data distribution identical to the one used by LongCat-Flash. During mid-training, \longcat walks through 32K, 128K, 256K training phases, and is extrapolated to 1M with the power of YaRN. To enhance the long-context ability, we involve 500B tokens during the 32K and 128K stages, followed by 40B tokens during the 256K stage. Regarding long-context data composition, it basically follows below:
\begin{itemize}
    \item Reasoning-intensive data: enhancing the reasoning potential by expanding reasoning patterns;
    \item Agentic data: synthesizing large-scale agentic interactions by leveraging a vast array of task-oriented web content and thousands of model context protocol (MCP) servers;
    \item High-quality long-form data: integrating a diverse collection of long-form data, including curated, open-source, and synthetic books and textbooks;
    \item Repository-level code: integrating an extensive collection of full-repository codebases to enhance the capacity for solving real-world, cross-file programming challenges.
\end{itemize}
By strategically diversifying the data mixture, this composition ensures high-fidelity of data.

\paragraph{Post-training}

To quickly validate \longcat, we adopt a lightweight post-training pipeline. Specifically, we perform supervised fine-tuning (SFT) using a data distribution identical to that of LongCat-Flash~\citep{DBLP:journals/corr/abs-2509-01322}, but with only 50\% of its original volume. To maintain performance, the dataset was carefully curated to span critical domains, including instruction following, mathematics, coding, agentic tasks, and general knowledge. Subsequently, to align with human preferences and optimize model behavior, we employed Direct Preference Optimization (DPO)~\citep{DBLP:conf/nips/RafailovSMMEF23} alongside Reinforcement Fine-Tuning (RFT)~\citep{openai2024rft}. Compared to large-scale reinforcement learning (RL) approaches, our strategy achieves competitive performance while consuming minimal computational resources.

\section{Evaluation}

\paragraph{Effectiveness}

We first evaluate the base \longcat-Base. It is demonstrated in Table~\ref{tab:effectiveness_base} that LoZA would not degrade performance. Namely, after mid-training with sparsity, \longcat-Base remains comparable to LongCat-Flash-Base.

\begin{table}[t]
    \centering
    \caption{The effectiveness of \longcat-Base.}
    \begin{tabular}{lcccccc}
    \toprule
        \textbf{Method} & \textbf{MMLU-Pro} & \textbf{GPQA} & \textbf{BBH} & \textbf{GSM8K} & \textbf{HumanEval+} & \textbf{LongEval} \\
    \midrule
        LongCat-Flash-Base & 70.0 & 51.2 & 81.0 & 94.2 & 66.6 & 95.7 \\
        \longcat-Base & 69.9 & 54.6 & 81.6 & 93.8 & 67.1 & 99.3 \\
    \bottomrule
    \end{tabular}
    \label{tab:effectiveness_base}
\end{table}

\begin{table}[t]
    \centering
    \caption{The effectiveness of \longcat-Chat. The comparison results are derived on a diverse range of benchmarks covering an array of domains. The hybrid thinking model with \textsuperscript{\dag} is evaluated in chat mode for a fair comparison. The best result at each row is \textbf{boldfaced}.}
    \begin{tabular}{lc|ccc|c}
    \toprule
        & \textbf{Benchmark\textsubscript{metric}} & \makecell[c]{\textbf{GLM}\\\textbf{4.6\textsuperscript{\dag}}}& \makecell[c]{\textbf{DeepSeek}\\\textbf{V3.2\textsuperscript{\dag}}} & \makecell[c]{\textbf{LongCat-Flash}\\\textbf{Chat}} & \makecell[c]{\textbf{LongCat-Flash}\\\textbf{Exp-Chat}} \\
    \midrule
        \multirow{5}{*}{General} & MMLU\textsubscript{Acc} & 90.7 & \textbf{91.1} & 89.7 & 89.6 \\
        & CEval\textsubscript{Acc} & 89.6 & 89.6 & \textbf{90.4} & 89.9 \\
        & CMMLU\textsubscript{Acc} & \textbf{88.4} & 87.3 & 84.3 & 87.5 \\
        & IFEval\textsubscript{Acc} & 87.8 & 88.4 & \textbf{89.7} & 88.0 \\
        & GuideBench\textsubscript{Acc} & 83.8 & 87.0 & 81.0 & \textbf{90.4} \\
    \midrule
        \multirow{4}{*}{Math} & MATH-500\textsubscript{Acc} & 98.6 & 97.2 & 96.4 &\textbf{98.8} \\
        & AIME-24\textsubscript{Avg@32} & \textbf{83.7} & 73.9 & 70.4 & 83.1 \\
        & AIME-25\textsubscript{Avg@32} & \textbf{80.2} & 56.5 & 61.3 & 74.9 \\
        & BeyondAIME\textsubscript{Avg@10} & 52.8 & 42.2 & 43.0 & \textbf{59.2} \\
    \midrule
        \multirow{4}{*}{Code} 
        & Humaneval+\textsubscript{Pass@1} & \textbf{92.7} & 89.0 & 88.4 & 87.2 \\
        & MBPP+\textsubscript{Pass@1} & \textbf{83.6} & 79.9 & 79.6 & 79.1 \\
        &LCB\textsubscript{2408-2505, Pass@1} & 56.4 & \textbf{59.5} & 48.0 & 56.6 \\
        &FullStackBench\textsubscript{Pass@1} & 61.0 & 62.5 & 61.8 & \textbf{64.1} \\
    \midrule
        \multirow{2}{*}{STEM} & MMLU-Pro\textsubscript{Acc} & 81.5 & \textbf{84.3} & 82.7 & 84.0 \\
        & GPQA-Diamond\textsubscript{Avg@16} & 74.8 & 75.3 & 73.2 & \textbf{75.6} \\
    \midrule
        \multirow{3}{*}{Agent} & SWE-Bench-Verified\textsubscript{Acc} & 68.0 & \textbf{72.1} & 60.4 & 63.2 \\
        & Terminal-Bench\textsubscript{Acc} & 40.5 & \textbf{45.0} & 39.5 & 42.5 \\
        & $\tau^2$-Bench\textsubscript{Acc} & 69.1 & 64.0 & 68.8 & \textbf{69.5} \\
    \midrule
        \multirow{2}{*}{Multilingual} & {MMMLU}\textsubscript{Acc} & \textbf{87.2} & 86.7 & 81.7 & 85.2 \\
        & {MGSM\textsubscript{Acc}} & 91.1 & \textbf{94.9} & 87.5 & 94.6 \\
    \midrule
        \multirow{4}{*}{Long Context} & LongBenchV2\textsubscript{Acc} & 51.5 & \textbf{54.1} & 38.2 & 53.5 \\
        & MRCR\textsubscript{Acc} & 42.1 & 37.1 & 34.4 & \textbf{59.7} \\
        & HELMET\textsubscript{Acc} & 64.6 & 59.5 & 59.1 & \textbf{64.7} \\
        & Longform-Writing\textsubscript{Acc} & 70.0 & \textbf{73.9} & 51.3 & 69.6 \\
    \bottomrule
    \end{tabular}
    \label{tab:effectiveness}
\end{table}

\begin{figure}[ht]
    \centering
    \begin{subfigure}[b]{0.49\textwidth}
        \centering
        \includegraphics[width=\textwidth]{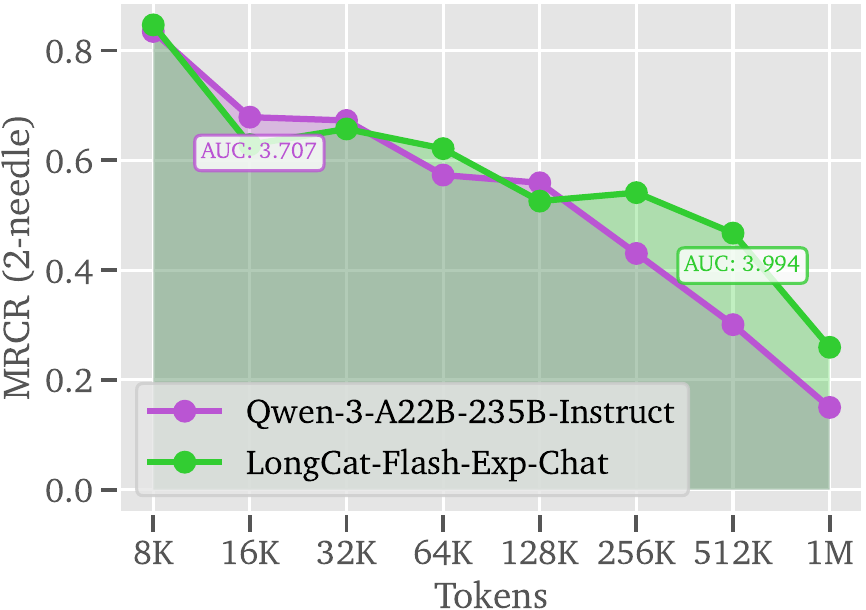}
        \caption{2-needle.}
        \label{fig:mrcr_2needle}
    \end{subfigure}
    \hfill
    \begin{subfigure}[b]{0.49\textwidth}
        \centering
        \includegraphics[width=\textwidth]{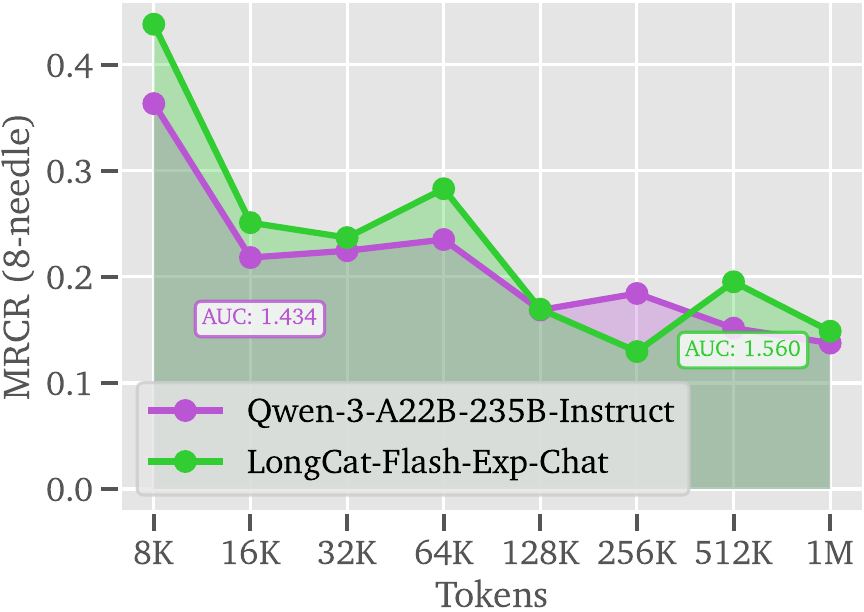}
        \caption{8-needle.}
        \label{fig:mrcr_8needle}
    \end{subfigure}
    \caption{The effectiveness of \longcat-Chat across different context lengths on MRCR. Qwen-3 is considered as a competitive baseline since it as well possesses the ability of handling 1M context. AUC: area under curve.}
    \label{fig:mrcr}
\end{figure}

We then evaluate the instructed \longcat-Chat on benchmarks of diverse domains and compare it against LongCat-Flash-Chat to examine the effectiveness of LoZA:
\begin{itemize}
    \item\textbf{General: }MMLU~\citep{hendrycks2021measuringmassivemultitasklanguage}, CEval~\citep{huang2023ceval}, and CMMLU~\citep{li2023cmmlu} for general knowledge. IFEval~\citep{zhou2023ifeval} and GuideBench~\citep{diao-etal-2025-guidebench} for instruction following.
    \item\textbf{Math: }Olympiad-level mathematical benchmarks, including MATH-500 \citep{math500}, AIME-24 \citep{AIME24} and AIME-25 \citep{AIME25} (American Invitational Mathematics Examinations), and BeyondAIME \citep{bytedanceseed2025beyondaime}.
    \item\textbf{STEM: }MMLU-Pro~\citep{wang2024mmluprorobustchallengingmultitask} and GPQA-Diamond \citep{rein2024gpqa}.
    \item\textbf{Code: }Humaneval+~\citep{humanevalmbppplus}, MBPP+~\citep{humanevalmbppplus}, 
    LiveCodeBench (2024.08-2025.05)~\citep{jain2025livecodebench}, and FullStackBench~\citep{liu2024fullstackbenchevaluatingllms}.
    \item\textbf{Agent: } 
    SWE-Bench~\citep{jimenez2024swebench}, sourced from real GitHub issues for evaluating a model's ability to solve software engineering problems. Terminal-Bench~\citep{tbench_2025}, for evaluating a model's agentic ability in real terminal environments.
    $\tau^2$-Bench~\citep{barres2025tau2}, a Tool-Augmented Reasoning benchmark.
    \item\textbf{Multilingual:\footnote{We report the average performance of eight widely-used languages, i.e., Arabic, French, German, Spanish, Portuguese, Indonesian, Japanese, and Korean.} } MMMLU~\citep{hendrycks2021measuringmassivemultitasklanguage} and MGSM~\citep{shi2022languagemodelsmultilingualchainofthought}. 
    \item\textbf{Long Context}: LongBenchV2~\citep{bai-etal-2025-longbench}, MRCR~\citep{vodrahalli2024michelangelolongcontextevaluations}, and HELMET~\citep{yen2025helmetevaluatelongcontextlanguage} for evaluating long-context understanding, and Longform-Writing~\citep{paech2025longform} for evaluating long text generation.
\end{itemize}

As shown in Table~\ref{tab:effectiveness}, LoZA would not compromise quality for speed. On the concerned benchmarks, \longcat-Chat exhibit competitive performance with LongCat-Flash-Chat. Concretely, \longcat excels LongCat-Flash-Chat on long-context benchmarks, largely due to the extended context length. \longcat-Chat also stands at the same line with other competitors such like GLM-4.6 concerning chat mode. That is, \longcat-Chat obtains similar number of \textbf{the best-performing ones} to that of GLM-4.6.

We also provide a micro-benchmarking of \longcat-Chat across different context lengths versus Qwen-3, which also possesses the ability of handling 1M context. In Figure~\ref{fig:mrcr}, we could clearly see that \longcat-Chat outweighs Qwen-3 on some context lengths and overall surpasses Qwen-3 in terms of AUC.\footnote{This metric for long-context evaluation is originally proposed in \url{https://contextarena.ai/}.} This implies that LoZA combined with YaRN could efficiently pave the way for context scale of 1M.

\paragraph{Efficiency}

We draw respectively the decode cost of SSA against full attention and the end-to-end timeline of \longcat against LongCat-Flash, to showcase how \longcat overwhelms LongCat-Flash in real-world serving.

\begin{figure}[t]
    \centering
    \begin{subfigure}[b]{0.82\textwidth}
        \centering
        \includegraphics[width=\textwidth]{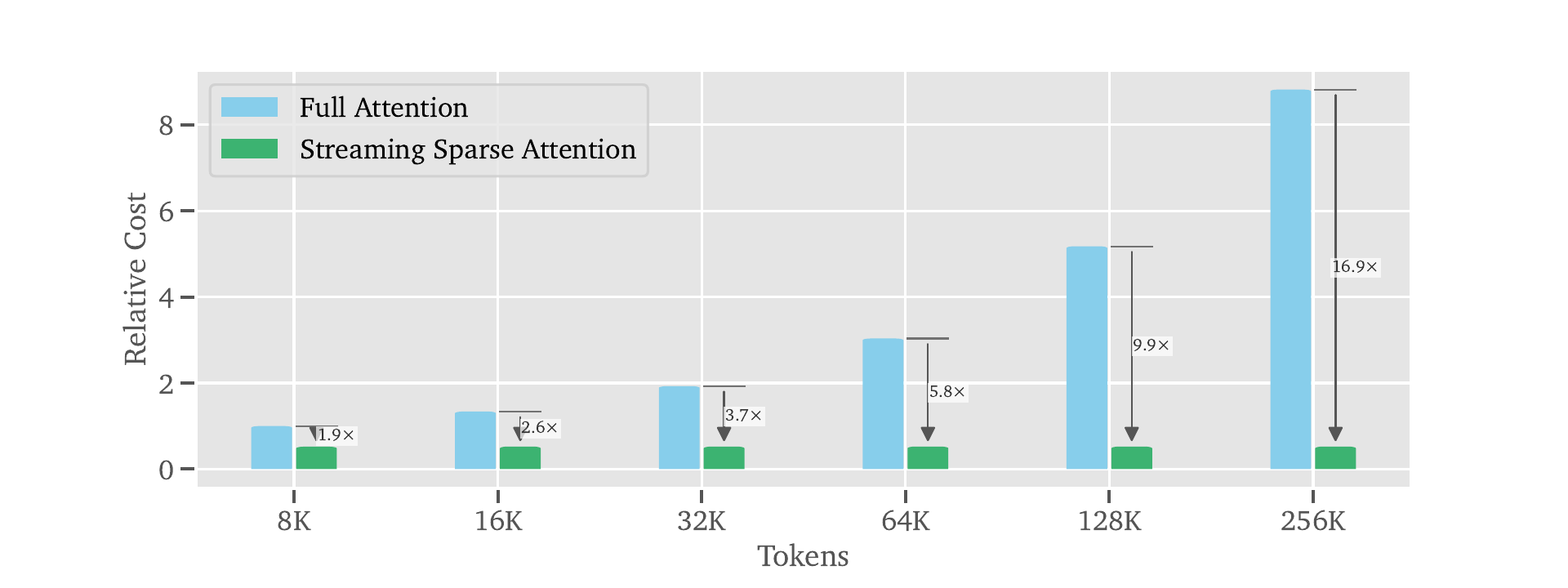}
        \caption{Decode kernel.}
        \label{fig:efficiency_op}
    \end{subfigure}
    \\
    \begin{subfigure}[b]{0.49\textwidth}
        \centering
        \includegraphics[width=\textwidth]{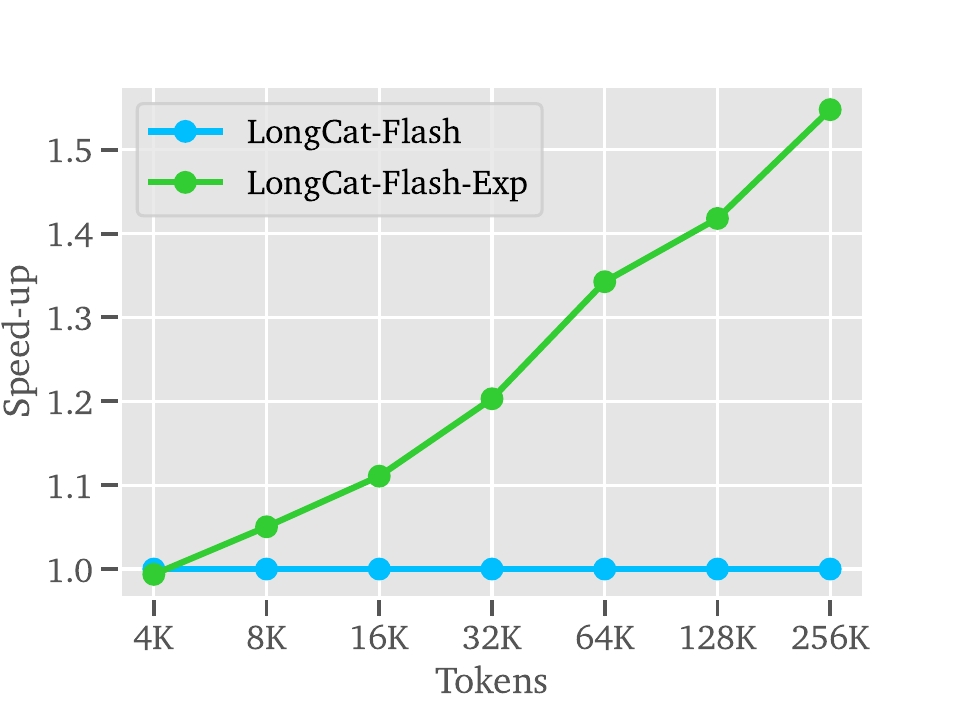}
        \caption{Prefill.}
        \label{fig:efficiency_prefill}
    \end{subfigure}
    \hfill
    \begin{subfigure}[b]{0.49\textwidth}
        \centering
        \includegraphics[width=\textwidth]{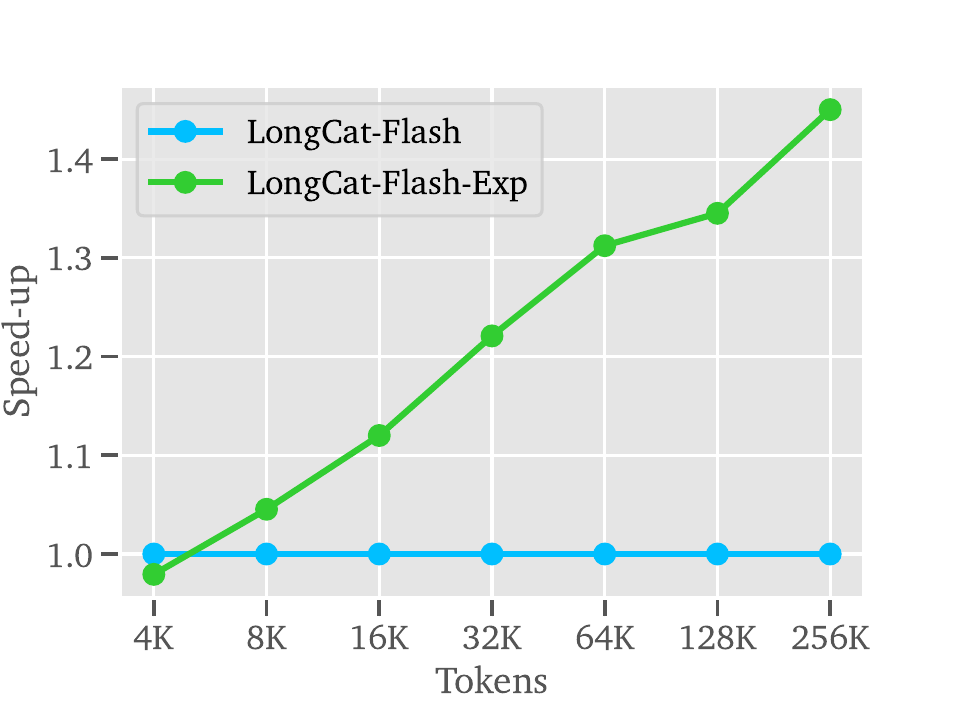}
        \caption{Decode.}
        \label{fig:efficiency_decode}
    \end{subfigure}
    \caption{The efficiency of LoZA. The relative cost and speed-up are practically measured on inference clusters.}
    \label{fig:efficiency}
\end{figure}

Since LoZA enables 50\% sparsity in \longcat, the compute brought by attention should be ideally reduced by a factor of 2. For long-context circumstances where attention dominates the compute, the efficiency could be maximally lifted to 2 times of the original. Promoted by our efforts in kernel and engine customizations, in Figure~\ref{fig:efficiency}, streaming sparse attention kernel could minimally use 90\% less cost in decode compared to full attention kernel (i.e., FlashMLA~\citep{DBLP:journals/corr/abs-2506-01969}) for a context of 128K tokens. Meanwhile, in end-to-end benchmarking, \longcat realizes more than 50\% speed-up in prefill and saves over 30\% cost in decode for a context of 256K tokens. 

\section{Conclusion}

We present LoZA, a sparse attention algorithm built upon MLA, that is universally applicable to full-attention LMs and based off LongCat-Flash results in \longcat. During mid-training, LoZA transforms LongCat-Flash via calibration, sparsification, and training. It is worthy of mentioning that the process essentially follows the pace of \emph{lottery tickets hypothesis}, providing adequate theoretical guarantee for LoZA. With specialized design efforts, LoZA realizes principal speed-ups in both prefill and decode phases. This enables efficient long-term reasoning and long-horizon agentic capabilities, thereby making context-native (i.e., context as memory) applications viable. LoZA might also broadly impact related work that is aims at improving attention, especially for that intends to transform the MLA to sparse one. We would extremely like to invite the community to embed LoZA into any other open-source LMs that use MLA, and perhaps large multi-modal models~\citep{Li2025OneCAT}.

\section*{Acknowledgement}

We sincerely thank the infrastructure team and evaluation team of LongCat for their constructive feedback and promptly support.

\bibliographystyle{unsrtnat}
\bibliography{ref}



\end{document}